\documentclass{article}

\usepackage{microtype}
\usepackage{graphicx}
\usepackage{subfigure}
\usepackage{booktabs} 
\usepackage[utf8]{inputenc}
\usepackage[T1]{fontenc}

\usepackage{hyperref}

\usepackage[accepted]{icml2019}

\icmltitlerunning{Crowdsourcing Insect Observations}

\begin{document}

\twocolumn[
\icmltitle{InsectUp: Crowdsourcing Insect Observations \\ to Assess Demographic Shifts and Improve Classification}

\begin{icmlauthorlist}
\icmlauthor{Léonard Boussioux}{mit,mila,ecole}
\icmlauthor{Tomás Giro-Larraz}{ecole,epfl}
\icmlauthor{Charles Guille-Escuret}{mila}
\icmlauthor{Mehdi Cherti}{cnrs,univ,mines}
\icmlauthor{Balázs Kégl}{cnrs,univ}
\end{icmlauthorlist}

\icmlaffiliation{mit}{Massachussets Institute of Technology, Cambridge, MA, USA}
\icmlaffiliation{mila}{Montreal Institute for Learning Algorithms, Montréal, Canada}
\icmlaffiliation{ecole}{Ecole CentraleSupélec, Gif-sur-Yvette, France}
\icmlaffiliation{epfl}{Ecole Polytechnique Fédérale de Lausanne, Lausanne, Suisse}
\icmlaffiliation{cnrs}{CNRS, Orsay, France}
\icmlaffiliation{univ}{Université Paris-Saclay, Orsay, France}
\icmlaffiliation{mines}{Mines ParisTech, Paris, France}

\icmlcorrespondingauthor{Léonard Boussioux}{leobix@mit.edu}

\icmlkeywords{Machine Learning, Insects, Entomology, Classification, Deep Learning, Computer Vision}

\vskip 0.3in ]
\printAffiliationsAndNotice{}




\begin{abstract}
Insects play such a crucial role in ecosystems that a shift in demography of just a few species can have devastating consequences at environmental, social and economic levels. Despite this, evaluation of insect demography is strongly limited by the difficulty of collecting census data at sufficient scale. We propose a method to gather and leverage observations from bystanders, hikers, and entomology enthusiasts in order to provide researchers with data that could significantly help anticipate and identify environmental threats. Finally, we show that there is indeed interest on both sides for such collaboration.
\end{abstract}

\section{Introduction}
It is estimated that 90\% of all animal life forms on Earth are insects \cite{2,1} with a ratio of 200 million specimens per human being \cite{5} at any time, and that there exist 6 to 10 million different species \cite{3}, of which we have only identified 900,000. With such numbers, it is tremendously difficult for entomologists to classify insect species, estimate current population distribution and detect their shifts. Yet, habitat loss, intensification of agricultural practices, urbanization and environmental changes result in insect decline \cite{10,11} and demographic shifts with dramatic consequences: 

\begin{figure}[ht]
\vskip 0.2in
\begin{center}
\centerline{\includegraphics[width=\columnwidth]{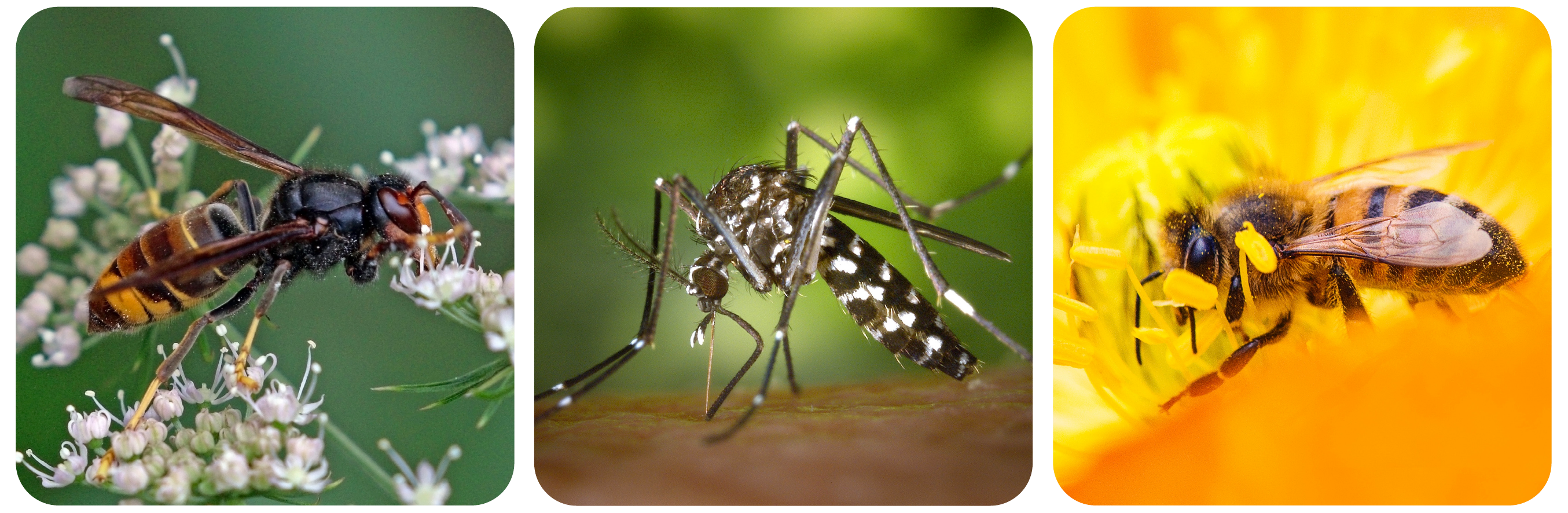}}
\caption{Asian hornet, tiger mosquito and honey bee.}
\label{tiger-mosquito}
\end{center}
\vskip -0.2in
\end{figure}

\textbf{Honey bees (Genus \textit{Apis})} have been disappearing at alarming rates in the first decade of the 21st century, due to a phenomenon called Colony Collapse Disorder. The causes are still unclear but are suspected to be a mix of environmental perturbations \cite{12}. Because agriculture depends heavily on bees and other pollinators, monitoring these populations is critical to anticipating and preventing the disasters that would follow large population declines.

\textbf{The Tiger Mosquito (\textit{Aedes albopictus})} is a widespread invasive species in the world. Its outstanding adaptation ability being boosted by global warming, it has outcompeted concurrent species on several continents. It is a vector for many dangerous diseases (yellow fever, dengue fever, Chikungunya fever and Usutu virus), making it a serious health concern worldwide. 

\textbf{The Asian Hornet (\textit{Vespa velutina})} is another invasive species \cite{4} with a population quickly increasing in Europe and particularly in France. It was introduced to this new habitat by humans and is a major concern due to several hospitalizations related to their stings and their aggressiveness toward local fauna, effectively destabilizing local ecosystems. 

Besides the direct impact of some demographic shifts, insect population distributions can also be used as powerful indicators of the ecosystem perturbations such as global warming, destruction of habitat, pesticides and pollution to justify environment protection policies and measure their effectiveness \cite{17}. \\
Since the scale of the problem makes it intractable for classical census methods, we propose an approach based on volunteer contributions from amateurs and bystanders, who benefit in return from insect identification tools, playful features and the appeal of contributing to crucial statistical studies.

\section{Related Work}

\subsection{Citizen Science}

The implication of the general public is gaining momentum as a method to efficiently collect a significant amount of data and annotations. These community-contributed data obtained on principles of volunteerism and public engagement are being increasingly used by scientific professionals in a variety of domains (e.g., \cite{cit2, cit3, comp, cit1, cit0}), and have been shown to expand biodiversity research \cite{citizen2, citizen3, citizen1}. This surge in the utility of citizen science data to biodiversity research is due largely to the development of web platforms such as eBird (www.ebird.org \cite{SULLIVAN20092282}), eButterfly (www.e-butterfly.org \cite{ebutterfly}), and iNaturalist (www.inaturalist.org) that curate and host the data at broad geographic scales. However, more specialized and localized citizen science projects can be valuable for specific taxonomic groups, and lead to new discoveries. For example, through the Natural History Museum of Los Angeles’ BioSCAN project, 43 new species of phorid flies were discovered in the Los Angeles basin \cite{bioscan1}, as well as highly unexpected drosophilid flies and parasitoid wasps \cite{bioscan3, bioscan2}.

Citizen science is particularly useful regarding the scale and scope of biodiversity surveys. Data collection can be scaled up to cover previously under-explored locations, as in the case of many urban environments. However, scientists still need to engage with the volunteers to ensure high data quality, which can also be improved by machine-learning-based approaches.

The power of massive online citizen science programs lies in the strength and diversity of its participants and stakeholders. Anyone with an interest in insects can participate -- new enthusiasts, backyard gardeners, curious bystanders, and seasoned experts alike. As more participants submit data, a collaborative environment will become the norm between insect enthusiasts, scientists, and conservationists. 

\subsection{AI-Assisted Crowd-sourcing Approaches}

We identify in particular two successful approaches relying on crowd-sourcing observations of living beings encouraged by a machine learning identification tool : \href{https://identify.plantnet.org/}{Pl@ntnet}, a mobile application for plant observation and identification, and \href{https://www.inaturalist.org/}{iNaturalist}, a platform encompassing all living beings. With roughly 500,000 users having made more than 18,000,000 observations over 210,000 species, it is a very popular tool that has proven public interest in this area. \\
However, the extremely large size of iNaturalist's scope has some drawbacks when dealing with insects specifically. Pictures that include both insects and plants may be difficult to identify due to their ambiguous nature, and because of how small the differences between species can be, insect manual identification sometimes requires specific expertise. \\
Moreover, entomologists have expressed their strong interest for observation protocols specifically designed for statistical studies of insects (see section 3.4), which can be greatly facilitated by a dedicated tool.

\section{Method}
\subsection{The Mobile Application}

Similarly to the \textit{Galaxy Zoo project} \cite{6}, which proposed astronomy amateurs to help to categorize galaxies and resulted in 40 million handmade classifications within 6 months, we seek to leverage the contributions made by users to help entomologists in their large scale tasks.

This is made through a mobile phone application called \textit{Anonymised}, which offers to identify insects on pictures taken by users. It also provides educational information on the identified insects and local fauna. The identified pictures can be used in turn as observations, to be used as demographic data along with picture location.

Additionally, the application helps users to identify insects on newly collected pictures. Along with a verification process to limit erroneous annotations, it allows a continuous enrichment of the dataset with new samples and species, effectively improving its own classification capabilities. 

\subsection{Classification Data}

Initial dataset is provided by the French Photographic Survey of Flower Visitors (SPIPOLL) \cite{13}, a project sponsored by the French National Museum of Natural History and Office for Insects and their Environment, of which a few examples are presented in Figure 2. 
It contains 145k crowd-sourced labeled observations over 403 species of insects. While the number of species is low, it is due to the combination of biological and geographical constraints: it only considers pollinating insects in mainland France. As more data is collected and manually annotated, the model can be retrained to increase its number of recognized species and accuracy. 

\begin{figure}[ht]
\vskip 0.2in
\begin{center}
\centerline{\includegraphics[width=\columnwidth]{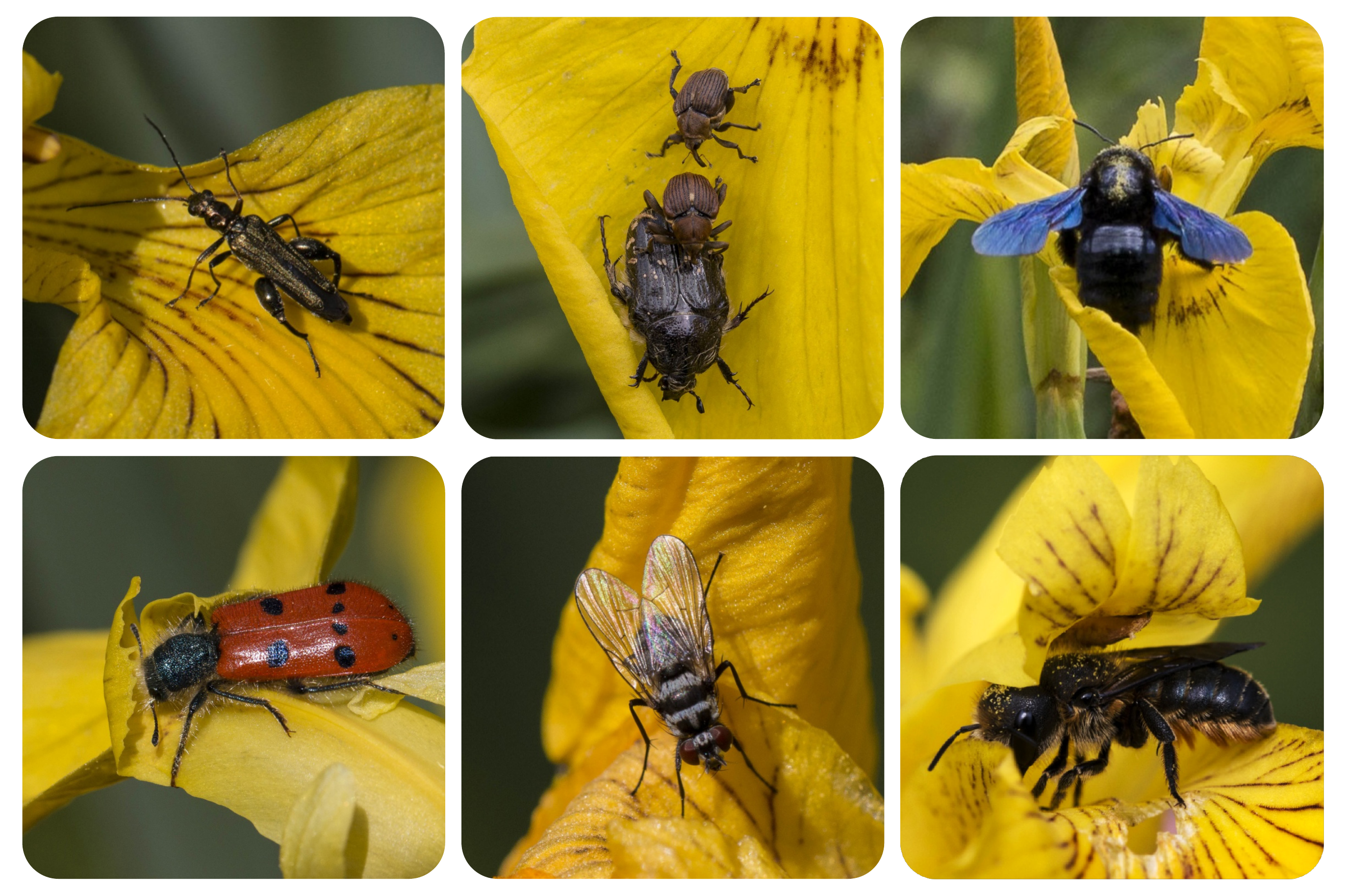}}
\caption{Example of SPIPOLL pictures obtained by capturing all visitors of an iris flower during 20 minutes.}
\label{spipoll}
\end{center}
\vskip -0.2in
\end{figure}

\subsection{Classification Model}

We rely on the RAMP framework \cite{7}, a platform developed for building transparent pipelines and easily reproducible results, and use as a model ResNet152 \cite{resnet}, pretrained on ImageNet \cite{14} then fine-tuned on our dataset. For training, we used a batch size of 8 and trained the model for 8 epochs. Each image was first center-cropped then resized to $224 \times 224$. We used standard SGD with an initial learning rate of 0.01. We divide the learning rate by 10 in the 4th epoch and the 7th epoch. To avoid overfitting, we used several data augmentation techniques such as horizontal and vertical flipping, random perturbations of contrast, random shift, random scale and random rotation. 
We obtain an 84\% top-1 accuracy on the SPIPOLL test set ($\sim$ 70k images).

\subsection{Interest from the Research Community and Partnerships}
The SPIPOLL project was originally launched as an effort from French entomologists to encourage citizen scientists to gather data on pollinating insects, following a standardized protocol,  to assess macro-ecological changes in richness and composition of flower visitor communities. This protocol will typically involve taking pictures of all insect visitors of a flower for a certain duration, then manually upload, identify and annotate each of these pictures to the SPIPOLL website. Such procedure can be greatly facilitated by the automation of both upload and identification of the pictures, and improving insect identification performance could lead to full automation of the process through autonomous cameras, which would multiply current insect census capabilities. The French National Center for Scientific Research (CNRS) and the French National Museum of Natural History have expressed their strong interest for data collected through such protocols. \\
Furthermore, even without observation protocols or autonomous cameras, the generated data can help estimate some demographic metrics such as the geographic distribution of a given species or its fluctuation over time. It is not implausible either to make original observations of species in regions where they had not been observed before or in time periods when they were not known as present.

However, it is important to acknowledge that this data may not be used directly to numerically estimate populations: indeed, it is subject to strong observer bias. Species that are more commonly seen in homes and urban areas, that are easy to notice because of their size and color, or are easier to capture (e.g. because they don't fly) will, for example, be over-represented in the collected data.

Finally, we have partnered with the Paris-Saclay Center for Data Science which provides us with technical support, access to their servers, and Amazon AWS funding.

\subsection{Autonomous Cameras}

Moreover, we partnered with \textit{Tidzam} \cite{tidzam}, a research effort for wildlife detection, identification and geo-localization in natural environments led by MIT Media Lab.
\textit{Tidzam} is a Deep Learning framework designed for the Tidmarsh Wildlife Sanctuary, the site of the largest freshwater wetland restoration in Massachusetts, USA. Through the installation of cameras and the use of detection algorithms, the fauna and flora can be monitored 24/7 in real-time, leading to rich biodiversity data that can be used to assess the health of the Tidmarsh's ecological system or improve understanding of ecological mechanisms and interactions. The aim of the collaboration is to use insect observations collected by volunteers to design powerful detection and classification algorithms, showcasing one of the potential high-impact application of such citizen-collected data. 

\section{Results}

\subsection{Engagement from Bystanders and Entomology Amateurs}
The mobile application \textit{Anonymised} was launched in April 2018 on the Android Playstore only, without any marketing. This launch served as an alpha to test its features and interest from users. \\
Despite the lack of visibility and the very limited features proposed by the app at this time, it had an unexpected success with over 50,000 downloads and more than 8,000 monthly active users at its peak in the middle of the Summer (Figure 3).
\begin{figure}[ht]
\vskip 0.2in
\begin{center}
\centerline{\includegraphics[width=\columnwidth] {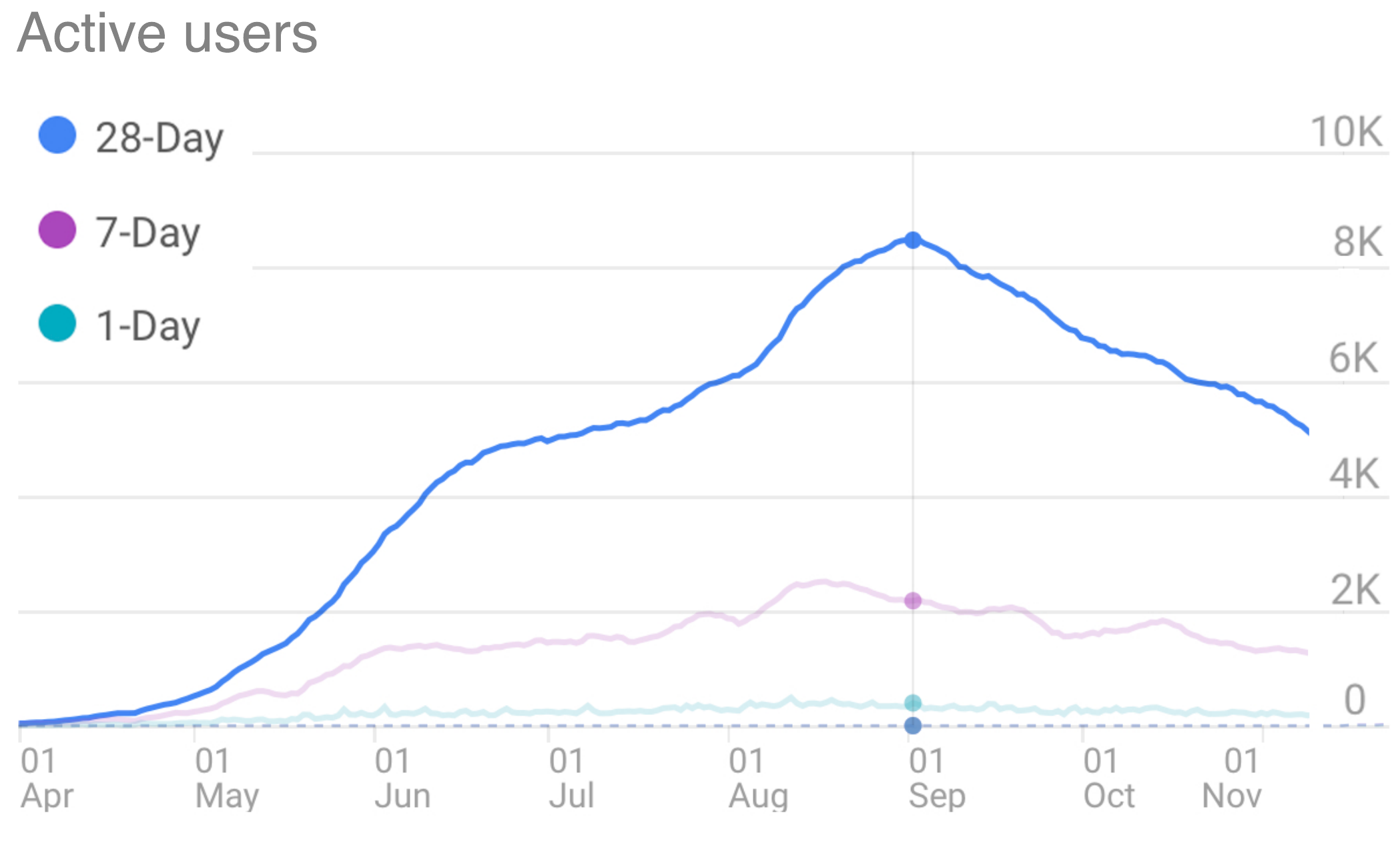}}
\caption{1-day, 7-day, and 28-day active users from April 2018 launch to November 2018, when the application was removed from the Playstore for refactoring.}
\label{users}
\end{center}
\vskip -0.2in
\end{figure}
Over the 7 months of the test, a total of more than 44,000 insect pictures have been uploaded by users (Figure 4), which shows that people are curious to identify insects and willing to participate actively. Additionally, there were many comments, identifications and discussions as the community progressively developed. \\
This proves the potential of this tool to reach a significant number of users, an essential condition to obtain data of sufficient scale.

\begin{figure}[ht]
\vskip 0.2in
\begin{center}
\centerline{\includegraphics[width=\columnwidth]{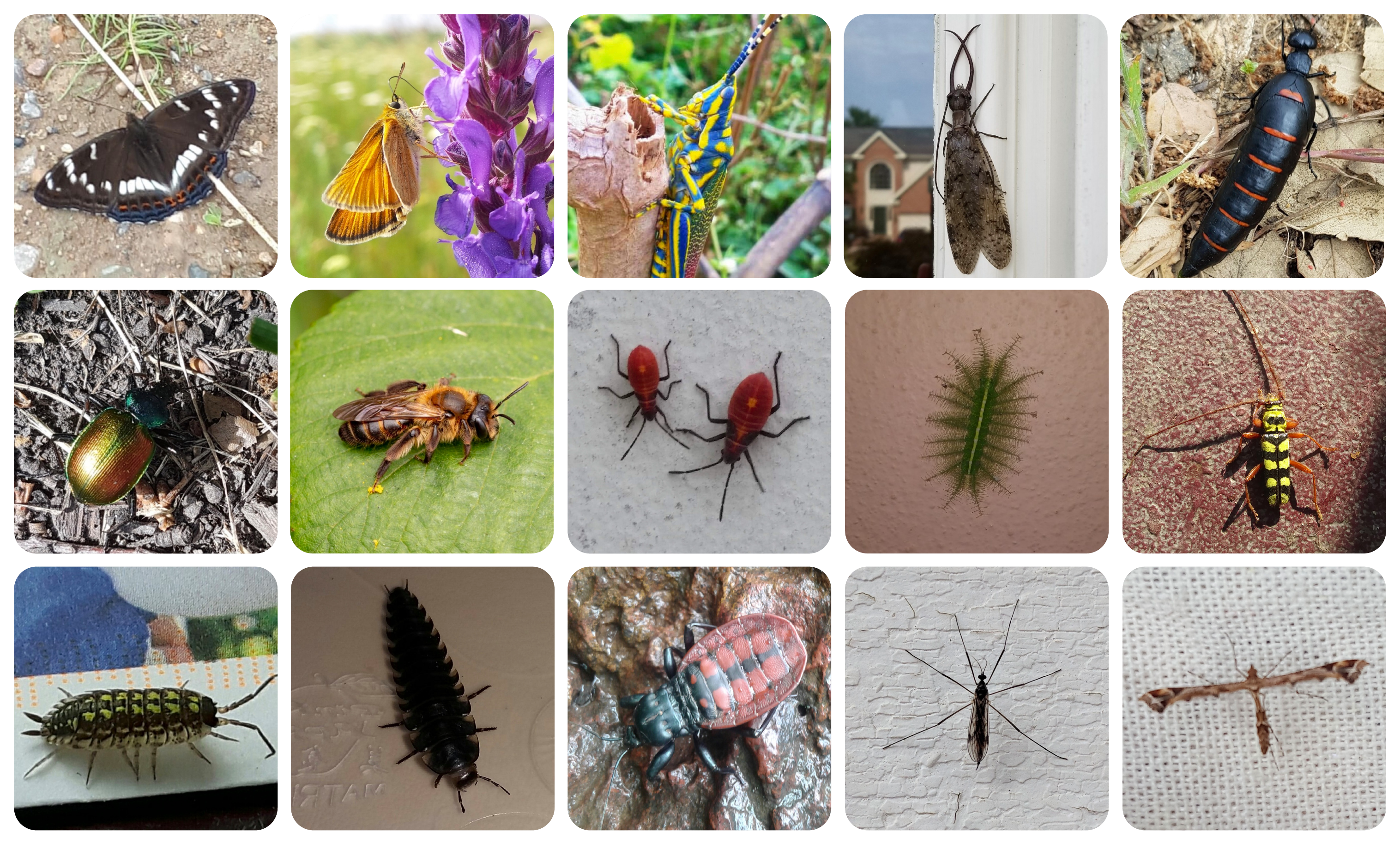}}
\caption{15 pictures taken by the users of our app and posted to our public gallery. Photo quality and insect species are very variable.}
\label{users-pics}
\end{center}
\vskip -0.2in
\end{figure}

\subsection{User Profile}
An interesting takeaway is that this project attracted a very wide community. Figure 5 shows that our application was downloaded all around the world, by users of all ages. It did not appeal only to entomology amateurs, but also to curious bystanders.
\begin{figure}[ht]
\vskip 0.2in
\begin{center}
\centerline{\includegraphics[width=\columnwidth]{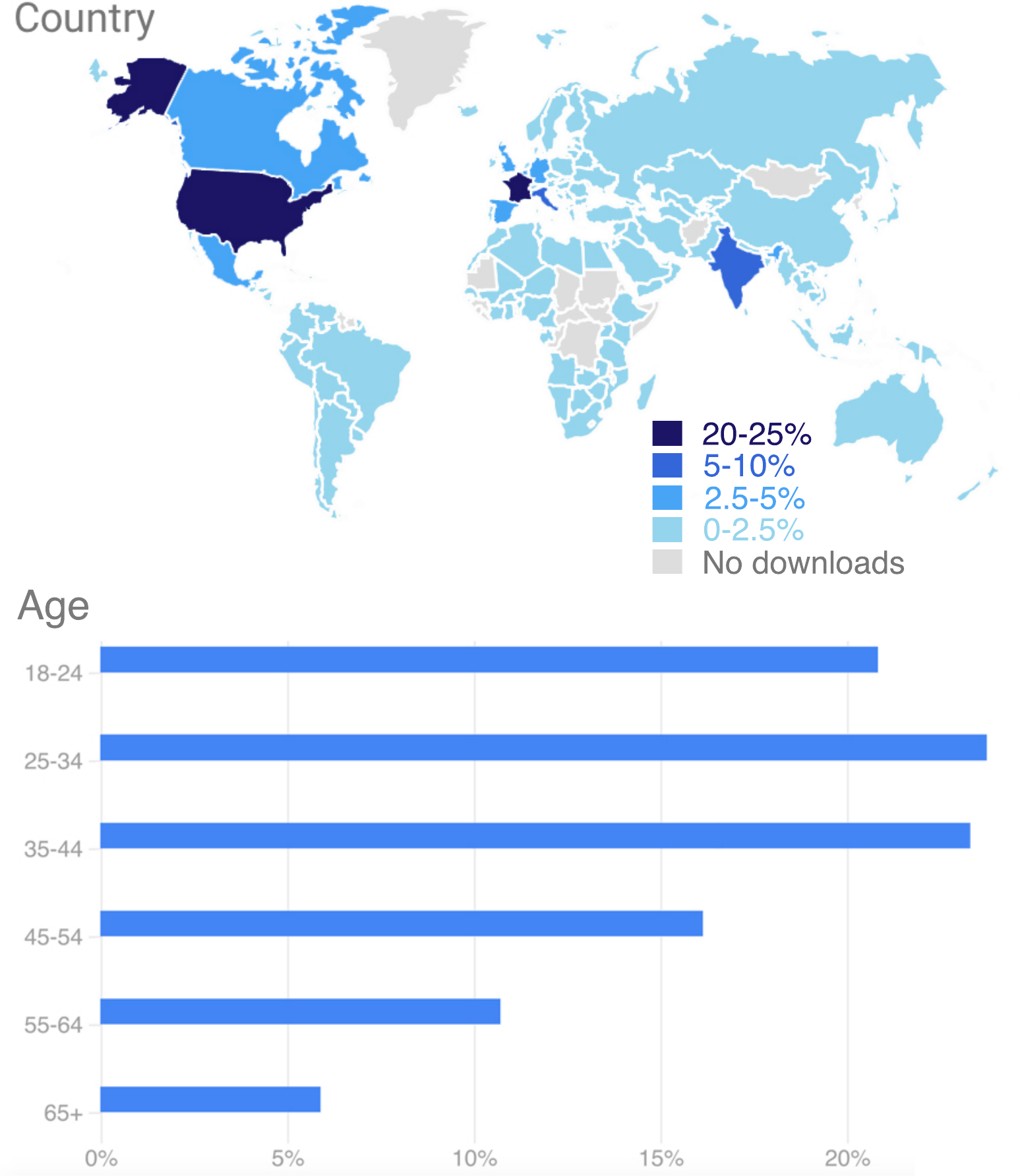}}
\caption{Age and geographic distribution of the application users.}
\label{demography}
\end{center}
\vskip -0.2in
\end{figure}

\section{Discussion}
While we believe to have shown the strong potential of this application for social good and insect conservation, a few technical challenges will need to be solved to fully take advantage of crowd-sourced observations. We present here 3 of these challenges.

\subsection{Coarse Classification for Tangent Cases}

With nearly a million insect species known on Earth, and many more not yet identified, it is to be expected that the amount of data available per species will be highly variable. Most species have never been classified by entomologists, and others are overwhelmingly represented due to their high population and easy observability. As the dataset grows in diversity, many species will only have a few annotated samples within the dataset, which can induce significant issues, especially on such a very large number of classes. The ability of the algorithm to quickly pick up on these new species to enrich the variety of its identifications is critical. 

While few-shot learning has recently made great progress \cite{16, 15}, these advances are largely targeted at \textit{n-shot}, \textit{k-way} meta-learning tasks, where the support dataset is part of the input. The size of our training data and the high variance in sample size between species call for a different solution. For example, the ability to perform a coarser classification for underrepresented species, such as identifying the genus or the family, could be an acceptable solution (see Figure 6). The algorithm should only classify at a level at which it is sufficiently confident. In cases where it is unable to classify further than the family or genus, the intervention of an human expert can refine the identification while significantly enriching the dataset.

\begin{figure}[ht]
\vskip 0.2in
\begin{center}
\centerline{\includegraphics[width=\columnwidth]{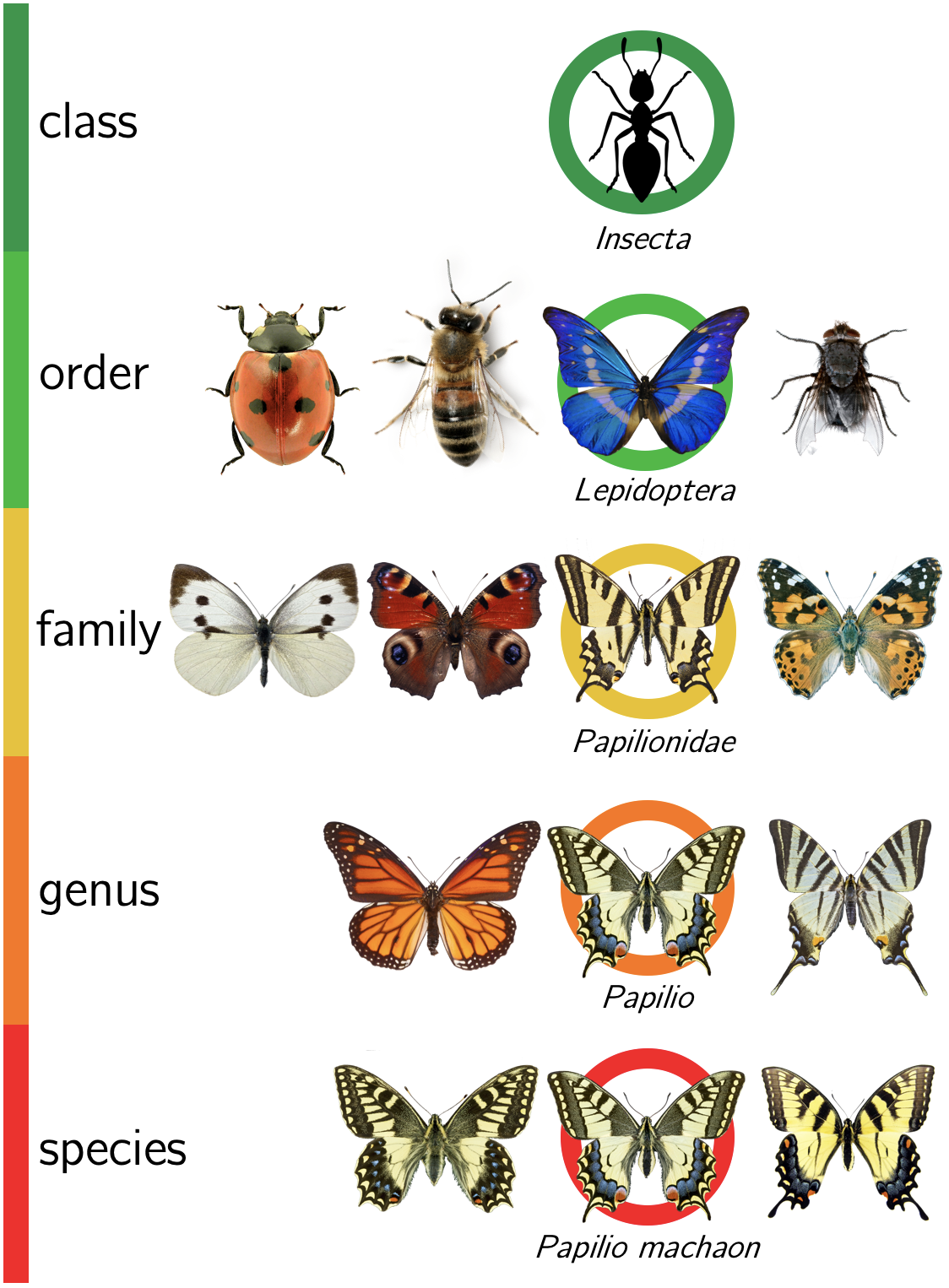}}
\caption{If confidence at species level is too low, we classify at genus level.}
\label{demography}
\end{center}
\vskip -0.2in
\end{figure}

\subsection{Rigorous Annotation Pipeline}

Since our approach is largely built around manual annotations from humans with different levels of expertise, a rigorous annotation pipeline needs to be designed to avoid erroneous identifications. This is especially difficult in the insect classification case because of the high level of similarity between some species. A good estimation of annotation confidence based on multiple identification suggestions and users identification history seems necessary, with possibly the intervention of a trained entomologist for disputed observations. While ensuring satisfying reliability of the labels at a large scale may be challenging, experience has shown there is an active community of entomologists willing to perform high-quality identifications if the interface is well designed \cite{13, ebutterfly}.

\subsection{False Observations}

Another issue is false observations, which can, for example, be produced by users taking pictures of themselves to see which insect they supposedly look like, a behavior which has been observed during the alpha phase of our mobile application. Additionally, the upload of pictures found online, either to test the algorithm performances or to purposely cause harm, can significantly corrupt gathered data.
Dealing with false observations will be an important challenge to successfully scale up this method.

While there are many other interesting directions to explore, we believe these 3 are some of the most critical for the large scale success of our approach.

\section{Conclusion}
We have presented the early development of a crowd-sourced insect observation tool, with the aim of providing entomologists high scale demographic data on a wide range of insects. We have shown evidence of the interest from both the scientific community and amateurs to collaborate in a platform that could improve our understanding and assessment of insect populations, and shown how such knowledge largely benefits society through its critical role in environmental protection.

While there is still a lot to do on this specific application, we believe similar approaches based on collaboration between citizen scientists and researchers could be successfully applied to many fields where gathering data is expensive and critical.

\section*{Acknowledgements}

We thank Alexandre Gramfort, Colin Fontaine, Grégoire Loïs, Romain Julliard, and the French National Museum of Natural History for sharing the dataset and valuable insights on their needs.
We also thank Rachel Tourneix, Mathilde Bryant, Jason Boussioux, Martin Kégl for their precious feedback on the mobile application and Louis Maestrati, Anne-Flore Baron, David Yu-Tung Hui, Baptiste Goujaud for their help in reviewing the paper.
We thank Paris Saclay Center of Data Science for their support, Linear Accelerator Laboratory (LAL) and Virtual Data for their hardware support.
This work was also supported by a grant from the \textit{Associations des Centraliens et des Supélec of Languedoc-Roussillon} and from \textit{La Recherche} and \textit{Sciences et Avenir}.

\bibliography{biblio}
\bibliographystyle{icml2019}

\end{document}